\documentclass{article}
\usepackage{times}
\usepackage{graphicx} 
\usepackage{subfigure}
\usepackage{stfloats}
\usepackage{natbib}
\usepackage{algorithm}
\usepackage{algorithmic}

\usepackage[accepted]{icml2014}
\icmltitlerunning{Distributed Representations of Sentences and Documents}

\begin{document}

\twocolumn[
\icmltitle{Distributed Representations of Sentences and Documents}

\icmlauthor{Quoc Le}{qvl@google.com}
\icmlauthor{Tomas Mikolov}{tmikolov@google.com}
\icmladdress{Google Inc,
            1600 Amphitheatre Parkway, Mountain View, CA 94043}

\icmlkeywords{}

\vskip 0.3in
]

\begin{abstract}
Many machine learning algorithms require the input to be represented
as a fixed-length feature vector. When it comes to texts, one of the
most common fixed-length features is bag-of-words. Despite their
popularity, bag-of-words features have two major weaknesses: they lose
the ordering of the words and they also ignore semantics of the
words. For example, ``powerful,'' ``strong'' and ``Paris'' are equally
distant.  In this paper, we propose \emph{Paragraph Vector}, an
unsupervised algorithm that learns fixed-length feature
representations from variable-length pieces of texts, such as
sentences, paragraphs, and documents.  Our algorithm represents each
document by a dense vector which is trained to predict words in the
document. Its construction gives our algorithm the potential to
overcome the weaknesses of bag-of-words models. Empirical results show
that Paragraph Vectors outperform bag-of-words models as well as
other techniques for text representations. Finally, we achieve new
state-of-the-art results on several text classification and sentiment
analysis tasks.
\end{abstract}

\section{Introduction}
\label{sec:intro}

Text classification and clustering play an important role in many
applications, e.g, document retrieval, web search, spam filtering. At
the heart of these applications is machine learning algorithms such as
logistic regression or K-means. These algorithms typically require the
text input to be represented as a fixed-length vector. Perhaps the
most common fixed-length vector representation for texts is the
bag-of-words or bag-of-n-grams~\cite{harris54} due to its simplicity,
efficiency and often surprising accuracy.

However, the bag-of-words (BOW) has many disadvantages.  The word
order is lost, and thus different sentences can have exactly the same
representation, as long as the same words are used. Even though
bag-of-n-grams considers the word order in short context, it suffers
from data sparsity and high dimensionality. Bag-of-words and
bag-of-n-grams have very little sense about the semantics of the words
or more formally the distances between the words. This means that
words ``powerful,'' ``strong'' and ``Paris'' are equally distant
despite the fact that semantically, ``powerful'' should be closer to
``strong'' than ``Paris.''

In this paper, we propose \emph{Paragraph Vector}, an unsupervised
framework that learns continuous distributed vector representations
for pieces of texts. The texts can be of variable-length, ranging from
sentences to documents. The name Paragraph Vector is to emphasize the
fact that the method can be applied to variable-length pieces of
texts, anything from a phrase or sentence to a large document.

In our model, the vector representation is trained to be useful for
predicting words in a paragraph.  More precisely, we concatenate the
paragraph vector with several word vectors from a paragraph and
predict the following word in the given context. Both word vectors and
paragraph vectors are trained by the stochastic gradient descent and
backpropagation~\cite{rumelhart1986learning}. While paragraph vectors
are unique among paragraphs, the word vectors are shared. At
prediction time, the paragraph vectors are inferred by fixing the word
vectors and training the new paragraph vector until convergence.

Our technique is inspired by the recent work in learning vector
representations of words using neural
networks~\cite{bengio2006neural,collobert2008unified,mnih2008scalable,turian2010word,mikolov,phrases0}.
In their formulation, each word is represented by a vector which is
concatenated or averaged with other word vectors in a context, and the
resulting vector is used to predict other words in the context.  For
example, the neural network language model proposed
in~\cite{bengio2006neural} uses the concatenation of several previous
word vectors to form the input of a neural network, and tries to
predict the next word.
The outcome is that after the model is trained, the word vectors are
mapped into a vector space such that semantically similar words have
similar vector representations (e.g., ``strong'' is close to
``powerful'').

Following these successful techniques, researchers have tried to
extend the models to go beyond word level to achieve phrase-level or
sentence-level
representations~\cite{mitchell10,zanzotto10,yessenalina11,
  grefen13,phrases0}. For instance, a simple approach is using a
weighted average of all the words in the document. A more
sophisticated approach is combining the word vectors in an order given
by a parse tree of a sentence, using matrix-vector
operations~\cite{socher2011parsing}.  Both approaches have
weaknesses. The first approach, weighted averaging of word vectors,
loses the word order in the same way as the standard bag-of-words
models do. The second approach, using a parse tree to combine word
vectors, has been shown to work for only sentences because it relies
on parsing.

Paragraph Vector is capable of constructing representations of input
sequences of variable length. Unlike some of the previous approaches,
it is general and applicable to texts of any length: sentences,
paragraphs, and documents.  It does not require task-specific tuning
of the word weighting function nor does it rely on the parse trees.
Further in the paper, we will present experiments on several benchmark
datasets that demonstrate the advantages of Paragraph Vector. For
example, on sentiment analysis task, we achieve new state-of-the-art
results, better than complex methods, yielding a relative improvement
of more than 16\% in terms of error rate. On a text classification
task, our method convincingly beats bag-of-words models, giving a
relative improvement of about 30\%.

\section{Algorithms}
We start by discussing previous methods for learning word
vectors. These methods are the inspiration for our Paragraph Vector
methods.

\subsection{Learning Vector Representation of Words}

\begin{figure}[bht]
\includegraphics[width=0.8\columnwidth]{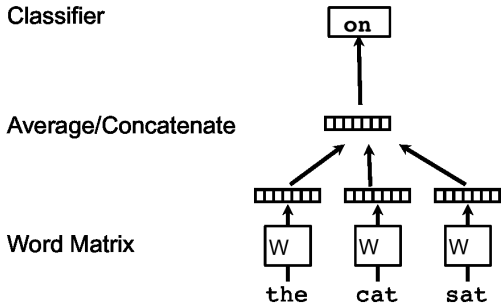}
\caption{A framework for learning word vectors. Context of three words
  (``the,'' ``cat,'' and ``sat'') is used to predict the fourth word
  (``on''). The input words are mapped to columns of the matrix $W$ to
  predict the output word.}
\label{fig:lm}
\end{figure}

This section introduces the concept of distributed vector
representation of words.  A well known framework for learning the word
vectors is shown in Figure~\ref{fig:lm}. The task is to predict a word
given the other words in a context.

In this framework, every word is mapped to a unique vector,
represented by a column in a matrix $W$. The column is indexed by
position of the word in the vocabulary. The concatenation or sum of
the vectors is then used as features for prediction of the next word
in a sentence.  

More formally, given a sequence of training words $w_1, w_2, w_3, ...,
w_T$, the objective of the word vector model is to maximize the
average log probability
\[
\frac{1}{T}\sum_{t=k}^{T-k} \log p (w_t | w_{t-k}, ..., w_{t+k})
\]

The prediction task is typically done via a multiclass
classifier, such as softmax. There, we have
\[
p (w_t | w_{t-k}, ..., w_{t+k}) = \frac{e^{y_{w_{t}}}}{\sum_i{e^{y_i}}}
\]
Each of $y_i$ is un-normalized log-probability for each output word $i$, computed as
\begin{equation}
y = b + U h(w_{t-k}, ..., w_{t+k}; W)
\label{eq:formal}
\end{equation}
where $U, b$ are the softmax parameters. $h$ is constructed by a
concatenation or average of word vectors extracted from $W$.

In practice, hierarchical
softmax~\cite{hsoft_first,mnih2008scalable,phrases0} is preferred to
softmax for fast training. In our work, the structure of the
hierarical softmax is a binary Huffman tree, where short codes are
assigned to frequent words. This is a good speedup trick because
common words are accessed quickly. This use of binary Huffman code for
the hierarchy is the same with~\cite{phrases0}.

The neural network based word vectors are usually trained using
stochastic gradient descent where the gradient is obtained via
backpropagation~\cite{rumelhart1986learning}. This type of models is
commonly known as neural language models~\cite{bengio2006neural}.  A
particular implementation of neural network based algorithm for
training the word vectors is available at {\tt
  code.google.com/p/word2vec/}~\cite{mikolov}.


After the training converges, words with similar meaning are mapped to
a similar position in the vector space. For example, ``powerful'' and
``strong'' are close to each other, whereas ``powerful'' and ``Paris''
are more distant. The difference between word vectors also carry
meaning. For example, the word vectors can be used to answer analogy
questions using simple vector algebra: ``King'' - ``man'' + ``woman''
= ``Queen''~\cite{mikolov2013naacl}.  It is also possible to learn a
linear matrix to translate words and phrases between
languages~\cite{MikolovLS13}.

These properties make word vectors attractive for many natural
language processing tasks such as language
modeling~\cite{bengio2006neural,mikolov2012}, natural language
understanding~\cite{collobert2008unified,Zhila}, statistical machine
translation~\cite{MikolovLS13,zou13}, image
understanding~\cite{Frome13} and relational
extraction~\cite{Socher13b}.

\subsection{Paragraph Vector: A distributed memory model}

Our approach for learning paragraph vectors is inspired by the methods
for learning the word vectors. The inspiration is that the word
vectors are asked to contribute to a prediction task about the next
word in the sentence. So despite the fact that the word vectors are
initialized randomly, they can eventually capture semantics as an
indirect result of the prediction task. We will use this idea in our
paragraph vectors in a similar manner. The paragraph vectors are also
asked to contribute to the prediction task of the next word given many
contexts sampled from the paragraph.

\begin{figure}[htb]
\includegraphics[width=\columnwidth]{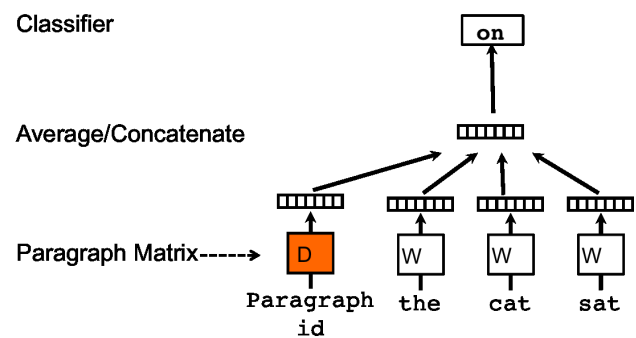}
\caption{A framework for learning paragraph vector. This framework is
  similar to the framework presented in Figure~\ref{fig:lm}; the only
  change is the additional paragraph token that is mapped to a vector
  via matrix $D$. In this model, the concatenation or average of this
  vector with a context of three words is used to predict the fourth
  word. The paragraph vector represents the missing information from
  the current context and can act as a memory of the topic of the
  paragraph.}
\label{fig:lm2}
\end{figure}

In our Paragraph Vector framework (see Figure~\ref{fig:lm2}), every
paragraph is mapped to a unique vector, represented by a column in
matrix $D$ and every word is also mapped to a unique vector,
represented by a column in matrix $W$. The paragraph vector and word
vectors are averaged or concatenated to predict the next word in a
context. In the experiments, we use concatenation as the method to
combine the vectors. 

More formally, the only change in this model compared to the word
vector framework is in equation \ref{eq:formal}, where $h$ is
constructed from $W$ and $D$.

The paragraph token can be thought of as another word. It acts as a
memory that remembers what is missing from the current context -- or
the topic of the paragraph. For this reason, we often call this model
the Distributed Memory Model of Paragraph Vectors (PV-DM). 

The contexts are fixed-length and sampled from a sliding window over
the paragraph. The paragraph vector is shared across all contexts
generated from the same paragraph but not across paragraphs. The word
vector matrix $W$, however, is shared across paragraphs. I.e., the
vector for ``powerful'' is the same for all paragraphs.

The paragraph vectors and word vectors are trained using stochastic
gradient descent and the gradient is obtained via backpropagation. At
every step of stochastic gradient descent, one can sample a
fixed-length context from a random paragraph, compute the error
gradient from the network in Figure~\ref{fig:lm2} and use the gradient
to update the parameters in our model.

At prediction time, one needs to perform an inference step to compute
the paragraph vector for a new paragraph. This is also obtained by
gradient descent. In this step, the parameters for the rest of the
model, the word vectors $W$ and the softmax weights, are fixed.


Suppose that there are $N$ paragraphs in the corpus, $M$ words in the
vocabulary, and we want to learn paragraph vectors such that each
paragraph is mapped to $p$ dimensions and each word is mapped to $q$
dimensions, then the model has the total of $N \times p + M\times q$
parameters (excluding the softmax parameters). Even though the number
of parameters can be large when $N$ is large, the updates during
training are typically sparse and thus efficient.

After being trained, the paragraph vectors can be used as features for
the paragraph (e.g., in lieu of or in addition to bag-of-words). We can
feed these features directly to conventional machine learning
techniques such as logistic regression, support vector machines or
K-means.

In summary, the algorithm itself has two key stages: 1) training to
get word vectors $W$, softmax weights $U,b$ and paragraph vectors $D$
on already seen paragraphs; and 2) \emph{``the inference stage''} to
get paragraph vectors $D$ for new paragraphs (never seen before) by
adding more columns in $D$ and gradient descending on $D$ while
holding $W,U,b$ fixed. We use $D$ to make a prediction about some
particular labels using a standard classifier, e.g., logistic
regression.


\paragraph{Advantages of paragraph vectors:}
An important advantage of paragraph vectors is that they are learned
from unlabeled data and thus can work well for tasks that do not have
enough labeled data.

Paragraph vectors also address some of the key weaknesses of
bag-of-words models. First, they inherit an important property of the
word vectors: the semantics of the words. In this space, ``powerful''
is closer to ``strong'' than to ``Paris.''  The second advantage of
the paragraph vectors is that they take into consideration the word
order, at least in a small context, in the same way that an n-gram
model with a large n would do. This is important, because the n-gram
model preserves a lot of information of the paragraph, including the
word order. That said, our model is perhaps better than a
bag-of-n-grams model because a bag of n-grams model would create a
very high-dimensional representation that tends to generalize poorly.


\subsection{Paragraph Vector without word ordering: Distributed bag of words}

\begin{figure}[htb]
\includegraphics[width=0.8\columnwidth]{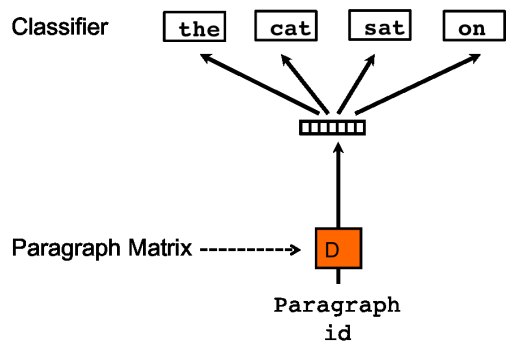}
\caption{Distributed Bag of Words version of paragraph vectors. In
  this version, the paragraph vector is trained to predict the words
  in a small window.}
\label{fig:lm3}
\end{figure}

The above method considers the concatenation of the paragraph vector
with the word vectors to predict the next word in a text
window. Another way is to ignore the context words in the input, but
force the model to predict words randomly sampled from the paragraph
in the output. In reality, what this means is that at each iteration
of stochastic gradient descent, we sample a text window, then sample a
random word from the text window and form a classification task given
the Paragraph Vector. This technique is shown in
Figure~\ref{fig:lm3}. We name this version the Distributed Bag of
Words version of Paragraph Vector (PV-DBOW), as opposed to Distributed
Memory version of Paragraph Vector (PV-DM) in previous section.

In addition to being conceptually simple, this model requires to store
less data. We only need to store the softmax weights as opposed to
both softmax weights and word vectors in the previous model. This
model is also similar to the Skip-gram model in word
vectors~\cite{phrases0}.

In our experiments, each paragraph vector is a combination of two
vectors: one learned by the standard paragraph vector with distributed
memory (PV-DM) and one learned by the paragraph vector with
distributed bag of words (PV-DBOW). PV-DM alone usually works well for
most tasks (with state-of-art performances), but its combination with
PV-DBOW is usually more consistent across many tasks that we try and
therefore strongly recommended.

\section{Experiments}
We perform experiments to better understand the behavior of the
paragraph vectors. To achieve this, we benchmark Paragraph Vector on
two text understanding problems that require fixed-length vector
representations of paragraphs: sentiment analysis and information
retrieval.

For sentiment analysis, we use two datasets: Stanford sentiment
treebank dataset~\cite{socher13} and IMDB
dataset~\cite{maas11}. Documents in these datasets differ
significantly in lengths: every example in Socher et
al.~\cite{socher13}'s dataset is a single sentence while every example
in Maas et al.~\cite{maas11}'s dataset consists of several sentences.

We also test our method on an information retrieval task, where the
goal is to decide if a document should be retrieved given a query.

\subsection{Sentiment Analysis with the Stanford Sentiment Treebank Dataset}
\paragraph{Dataset:}
This dataset was first proposed by~\cite{Pang+Lee:05a} and
subsequently extended by~\cite{socher13} as a benchmark for sentiment
analysis. It has 11855 sentences taken from the movie review site
Rotten Tomatoes. 


The dataset consists of three sets: 8544 sentences for training, 2210
sentences for test and 1101 sentences for validation (or development).

Every sentence in the dataset has a label which goes from very
negative to very positive in the scale from 0.0 to 1.0. The labels are
generated by human annotators using Amazon Mechanical Turk.

The dataset comes with detailed labels for sentences, and subphrases
in the same scale. To achieve this, Socher et al.~\cite{socher13} used
the Stanford Parser~\cite{klein03} to parse each sentence to
subphrases. The subphrases were then labeled by human annotators in
the same way as the sentences were labeled. In total, there are
239,232 labeled phrases in the dataset. The dataset can be downloaded
at: {\tt http://nlp.Stanford.edu/sentiment/}

\paragraph{Tasks and Baselines:}
In~\cite{socher13}, the authors propose two ways of
benchmarking. First, one could consider a 5-way \emph{fine-grained}
classification task where the labels are \{Very Negative, Negative,
Neutral, Positive, Very Positive\} or a 2-way \emph{coarse-grained}
classification task where the labels are \{Negative,
Positive\}. The other axis of variation is in terms of whether we
should label the entire sentence or all phrases in the sentence.
In this work we only consider labeling the full sentences.

Socher et al.~\cite{socher13} apply several methods to this dataset
and find that their Recursive Neural Tensor Network works much better
than bag-of-words model. It can be argued that this is because movie
reviews are often short and compositionality plays an important role
in deciding whether the review is positive or negative, as well as
similarity between words does given the rather tiny size of the
training set.

\paragraph{Experimental protocols:}
We follow the experimental protocols as described
in~\cite{socher13}. To make use of the available labeled data, in our
model, each subphrase is treated as an independent sentence and we
learn the representations for all the subphrases in the training set. 

After learning the vector representations for training sentences and
their subphrases, we feed them to a logistic regression to learn
a predictor of the movie rating.

At test time, we freeze the vector representation for each word, and
learn the representations for the sentences using gradient
descent. Once the vector representations for the test sentences are
learned, we feed them through the logistic regression to predict the
movie rating.

In our experiments, we cross validate the window size using the
validation set, and the optimal window size is 8. The vector presented
to the classifier is a concatenation of two vectors, one from PV-DBOW
and one from PV-DM. In PV-DBOW, the learned vector representations
have 400 dimensions. In PV-DM, the learned vector representations have
400 dimensions for both words and paragraphs. To predict the 8-th
word, we concatenate the paragraph vectors and 7 word vectors. Special
characters such as ,.!?  are treated as a normal word. If the
paragraph has less than 9 words, we pre-pad with a special NULL word
symbol.

\begin{table}[htb]
\caption{The performance of our method compared to other approaches on
  the Stanford Sentiment Treebank dataset. The error rates of other
  methods are reported in~\cite{socher13}.}
\label{tab:treebank}
\begin{center}
\begin{small}
\begin{tabular}{|l|r|r|}
\hline
Model & Error rate          & Error rate\\ 
      & (Positive/ & (Fine-\\ 
      & Negative)  &  grained)             \\\hline
Na\"{\i}ve Bayes    & 18.2 \%             & 59.0\% \\ 
\cite{socher13} &                    &         \\\hline
SVMs~\cite{socher13}   & 20.6\%              & 59.3\% \\ \hline
Bigram Na\"{\i}ve Bayes  & 16.9\%               & 58.1\%\\ 
\cite{socher13}     &                      &       \\\hline
Word Vector Averaging & 19.9\%             & 67.3\%\\ 
\cite{socher13}       &                    &        \\\hline 
Recursive Neural Network  & 17.6\%               & 56.8\%\\ 
\cite{socher13}          &                      &     \\\hline
Matrix Vector-RNN & 17.1\%             & 55.6\%\\ 
\cite{socher13}   &                    &      \\\hline
Recursive Neural Tensor Network & 14.6\%               & 54.3\%\\ 
\cite{socher13}                 &                      &  \\\hline
Paragraph Vector & {\bf 12.2\%}   & {\bf 51.3\%}\\
\hline
\end {tabular}
\end{small}
\end {center}
\end {table}

\paragraph{Results:} We report the error rates of different methods in
Table~\ref{tab:treebank}. The first highlight for this Table is that
bag-of-words or bag-of-n-grams models (NB, SVM, BiNB) perform
poorly. Simply averaging the word vectors (in a bag-of-words fashion)
does not improve the results. This is because bag-of-words models do
not consider how each sentence is composed (e.g., word ordering) and
therefore fail to recognize many sophisticated linguistic phenomena,
for instance sarcasm. The results also show that more advanced methods
(such as Recursive Neural Network~\cite{socher13}), which require
parsing and take into account the compositionality, perform much
better.

Our method performs better than all these baselines, e.g., recursive
networks, despite the fact that it does not require parsing. On the
coarse-grained classification task, our method has an absolute
improvement of 2.4\% in terms of error rates. This translates to 16\%
relative improvement.

\subsection{Beyond One Sentence: Sentiment Analysis with IMDB dataset}
Some of the previous techniques only work on sentences, but not
paragraphs/documents with several sentences.  For instance, Recursive
Neural Tensor Network~\cite{socher13} is based on the parsing over
each sentence and it is unclear how to combine the representations
over many sentences. Such techniques therefore are restricted to work
on sentences but not paragraphs or documents.

Our method does not require parsing, thus it can produce a
representation for a long document consisting of many sentences. This
advantage makes our method more general than some of the other approaches. The
following experiment on IMDB dataset demonstrates this advantage.

\paragraph{Dataset:}
The IMDB dataset was first proposed by Maas et al.~\cite{maas11} as a
benchmark for sentiment analysis. The dataset consists of 100,000
movie reviews taken from IMDB. One key aspect of this dataset is that
each movie review has several sentences. 




The 100,000 movie reviews are divided into three datasets: 25,000
labeled training instances, 25,000 labeled test instances and 50,000
unlabeled training instances. There are two types of labels: Positive
and Negative. These labels are balanced in both the training and
the test set. The dataset can be downloaded at {\tt
  http://ai.Stanford.edu/\newline~amaas/data/sentiment/index.html}

\paragraph{Experimental protocols:}
We learn the word vectors and paragraph vectors using 75,000 training
documents (25,000 labeled and 50,000 unlabeled instances). The
paragraph vectors for the 25,000 labeled instances are then fed
through a neural network with one hidden layer with 50 units and a
logistic classifier to learn to predict the sentiment.\footnote{In our
  experiments, the neural network did perform better than a linear
  logistic classifier in this task.}

At test time, given a test sentence, we again freeze the rest of the
network and learn the paragraph vectors for the test reviews by
gradient descent. Once the vectors are learned, we feed them through
the neural network to predict the sentiment of the reviews.

The hyperparameters of our paragraph vector model are selected in the
same manner as in the previous task. In particular, we cross validate
the window size, and the optimal window size is 10 words. The vector
presented to the classifier is a concatenation of two vectors, one
from PV-DBOW and one from PV-DM. In PV-DBOW, the learned vector
representations have 400 dimensions. In PV-DM, the learned vector
representations have 400 dimensions for both words and documents. To
predict the 10-th word, we concatenate the paragraph vectors and word
vectors. Special characters such as ,.!? are treated as a normal
word. If the document has less than 9 words, we pre-pad with a special
NULL word symbol.

\paragraph{Results:}
The results of Paragraph Vector and other baselines are reported in
Table~\ref{tab:imdb}. As can be seen from the Table, for long
documents, bag-of-words models perform quite well and it is difficult to
improve upon them using word vectors. The most significant improvement
happened in 2012 in the work of~\cite{dahl12} where they combine a
Restricted Boltzmann Machines model with bag-of-words. The combination
of two models yields an improvement approximately 1.5\% in terms of
error rates.

Another significant improvement comes from the work
of~\cite{wang12}. Among many variations they tried, NBSVM on bigram
features works the best and yields a considerable improvement of 2\% in
terms of the error rate.

The method described in this paper is the only
approach that goes significantly beyond the barrier of 10\% error rate.
It achieves
7.42\% which is another 1.3\% absolute improvement (or 15\% relative
improvement) over the best previous result of~\cite{wang12}.

\begin{table}[htb]
\caption{The performance of Paragraph Vector compared to other approaches on
  the IMDB dataset. The error rates of other methods are reported
  in~\cite{wang12}.}
\label{tab:imdb}
\begin{center}
\begin{tabular}{|l|r|} 
\hline
Model & Error rate \\ \hline
BoW (bnc)~\cite{maas11} & 12.20 \% \\ 
BoW (b$\Delta$t'c)~\cite{maas11} & 11.77\% \\ 
LDA~\cite{maas11} & 32.58\% \\ 
Full+BoW~\cite{maas11}& 11.67\%\\
Full+Unlabeled+BoW~\cite{maas11}& 11.11\% \\\hline 
WRRBM~\cite{dahl12}& 12.58\%\\
WRRBM + BoW (bnc)~\cite{dahl12}& 10.77\%\\\hline
MNB-uni~\cite{wang12} & 16.45\% \\ 
MNB-bi~\cite{wang12} & 13.41\%\\
SVM-uni~\cite{wang12} & 13.05\%\\
SVM-bi~\cite{wang12} & 10.84\%\\
NBSVM-uni~\cite{wang12} & 11.71\%\\
NBSVM-bi~\cite{wang12} & 8.78\%\\ \hline
Paragraph Vector & {\bf 7.42\%} \\
\hline
\end {tabular}
\end {center}
\end {table}

\subsection{Information Retrieval with Paragraph Vectors}
We turn our attention to an information retrieval task which requires
fixed-length representations of paragraphs.

Here, we have a dataset of paragraphs in the first 10 results returned
by a search engine given each of 1,000,000 most popular queries. Each
of these paragraphs is also known as a ``snippet'' which summarizes the
content of a web page and how a web page matches the query.

From such collection, we derive a new dataset to test vector
representations of paragraphs. For each query, we create a triplet of
paragraphs: the two paragraphs are results of the same query, whereas
the third paragraph is a randomly sampled paragraph from the rest of
the collection (returned as the result of a different query). Our goal
is to identify which of the three paragraphs are results of the same
query.  To achieve this, we will use paragraph vectors and compute the
distances the paragraphs. A better representation is one that achieves
a small distance for pairs of paragraphs of the same query and a larg
distance for pairs of paragraphs of different queries.

Here is a sample of three paragraphs, where the first paragraph should be
closer to the second paragraph than the third paragraph:

\begin{itemize}
\item {\bf Paragraph 1:}
calls from  ( 000 )  000 - 0000 .  3913 calls reported from this number .  according to 4  reports the identity of this caller is american airlines .  
\item {\bf Paragraph 2:}
do you want to find out who called you from +1 000 - 000 - 0000 ,  +1 0000000000  or  ( 000 )  000 - 0000 ?  see reports and share information you have about this caller
\item {\bf Paragraph 3:}
allina health clinic patients for your convenience ,  you can pay your allina health  clinic bill online .  pay your clinic bill now ,  question and answers...
\end{itemize}

The triplets are split into three sets: 80\% for training, 10\% for
validation, and 10\% for testing. Any method that requires learning
will be trained on the training set, while its hyperparameters will be
selected on the validation set.

We benchmark four methods to compute features for paragraphs:
bag-of-words, bag-of-bigrams, averaging word vectors and Paragraph
Vector. To improve bag-of-bigrams, we also learn a weighting matrix
such that the distance between the first two paragraphs is minimized
whereas the distance between the first and the third paragraph is
maximized (the weighting factor between the two losses is a
hyperparameter).

We record the number of times when each method produces smaller
distance for the first two paragraphs than the first and the third
paragraph. An error is made if a method does not produce that desirable
distance metric on a triplet of paragraphs.

The results of Paragraph Vector and other baselines are reported in
Table~\ref{tab:info}. In this task, we find that TF-IDF weighting
performs better than raw counts, and therefore we only report the
results of methods with TF-IDF weighting.

The results show that Paragraph Vector works well and gives a 32\%
relative improvement in terms of error rate. The fact that the
paragraph vector method significantly outperforms bag of words and
bigrams suggests that our proposed method is useful for capturing the
semantics of the input text.

\begin{table}[htb]
\caption{The performance of Paragraph Vector and bag-of-words models on the
  information retrieval task. ``Weighted Bag-of-bigrams'' is the
  method where we learn a linear matrix $W$ on TF-IDF bigram features
  that maximizes the distance between the first and the third paragraph
  and minimizes the distance between the first and the second
  paragraph.}
\label{tab:info}
\begin{center}
\begin{tabular}{|l|r|} 
\hline
Model & Error rate \\ \hline
Vector Averaging & 10.25\% \\
Bag-of-words & 8.10 \% \\ 
Bag-of-bigrams & 7.28 \% \\
Weighted Bag-of-bigrams & 5.67\% \\ 
Paragraph Vector & {\bf 3.82\%} \\
\hline
\end {tabular}
\end {center}
\end {table}

\subsection{Some further observations}
We perform further experiments to understand various aspects of the
models. Here's some observations
\begin{itemize}
\item PV-DM is consistently better than PV-DBOW. PV-DM alone can
  achieve results close to many results in this paper (see
  Table~\ref{tab:imdb}). For example, in IMDB, PV-DM only achieves
  7.63\%. The combination of PV-DM and PV-DBOW often work consistently
  better (7.42\% in IMDB) and therefore recommended.
\item Using concatenation in PV-DM is often better than sum. In IMDB,
  PV-DM with sum can only achieve 8.06\%. Perhaps, this is because the
  model loses the ordering information.
\item It's better to cross validate the window size. A good
  guess of window size in many applications is between 5 and 12. In
  IMDB, varying the window sizes between 5 and 12 causes the error
  rate to fluctuate 0.7\%.
\item Paragraph Vector can be expensive, but it can be done in
  parallel at test time. On average, our implementation takes 30
  minutes to compute the paragraph vectors of the IMDB test set, using
  a 16 core machine (25,000 documents, each document on average has
  230 words).
\end{itemize}

\section{Related Work}



Distributed representations for words were first proposed
in~\cite{rumelhart1986learning} and have become a successful paradigm,
especially for statistical language
modeling~\cite{elman,bengio2006neural,mikolov2012}. Word vectors have
been used in NLP applications such as word representation, named
entity recognition, word sense disambiguation, parsing, tagging and
machine
translation~\cite{collobert2008unified,turney10,turian2010word,collobert2011natural,socher2011parsing,huang2012improving,zou13}.

Representing phrases is a recent trend and received much
attention~\cite{mitchell10,zanzotto10,yessenalina11,
  grefen13,phrases0}. In this direction, autoencoder-style models have
also been used to model paragraphs~\cite{maas11,laro12,sri13}.

Distributed representations of phrases and sentences are also the
focus of Socher et al.~\cite{socher11,socher11b,socher13}. Their
methods typically require parsing and is shown to work for
sentence-level representations. And it is not obvious how to extend
their methods beyond single sentences. Their methods are also
supervised and thus require more labeled data to work well. Paragraph
Vector, in contrast, is mostly unsupervised and thus can work well
with less labeled data.

Our approach of computing the paragraph vectors via gradient descent
bears resemblance to a successful paradigm in computer
vision~\cite{perronnin07,perronnin10} known as Fisher
kernels~\cite{jaakkola99}. The basic construction of Fisher kernels is
the gradient vector over an unsupervised generative model.  



\section{Discussion}
We described Paragraph Vector, an unsupervised learning algorithm that
learns vector representations for variable-length pieces of texts such
as sentences and documents. The vector representations are learned to
predict the surrounding words in contexts sampled from the paragraph.


Our experiments on several text classification tasks such as Stanford
Treebank and IMDB sentiment analysis datasets show that the method is
competitive with state-of-the-art methods. The good performance
demonstrates the merits of Paragraph Vector in capturing the semantics of
paragraphs. In fact, paragraph vectors have the potential to overcome
many weaknesses of bag-of-words models.

Although the focus of this work is to represent texts, our method can
be applied to learn representations for sequential data. In non-text
domains where parsing is not available, we expect Paragraph Vector to
be a strong alternative to bag-of-words and bag-of-n-grams models.

\bibliography{translate}
\bibliographystyle{icml2014}

\end{document}